\documentclass[11pt]{article}

\usepackage[margin=1in]{geometry}
\usepackage{newtxtext}
\usepackage{microtype}
\usepackage{graphicx}
\usepackage{booktabs}
\usepackage{array}
\usepackage{multirow}
\usepackage{amsmath,amssymb}
\usepackage{xcolor}
\usepackage{hyperref}
\usepackage{url}
\usepackage{enumitem}
\usepackage{caption}
\usepackage{subcaption}
\usepackage{placeins}
\usepackage{flafter}
\usepackage{algorithm}
\usepackage{algpseudocode}
\usepackage{tikz}
\usetikzlibrary{positioning,arrows.meta,fit,backgrounds}

\hypersetup{colorlinks=true, linkcolor=blue, citecolor=blue, urlcolor=blue}

\newcommand{\successonce}{success-once}
\newcommand{\successend}{success-at-end}
\newcommand{\gad}{TAD}
\newcommand{\opd}{OPD}

\title{TAD: Distilling Vision-Language-Action Policies\\
from Teacher Trajectories Alone}

\author{%
  Yun Zhu$^{1}$ \quad Da Wang$^{2}$\thanks{Corresponding author.}  \\[1ex]
  $^{1}$Independent Researcher, China \\
  $^{2}$School of Computer and Information Technology, Shanxi University, Taiyuan, China \\[0.5ex]
  \texttt{wanda@sxu.edu.cn}
}

\begin{document}
\maketitle
\begin{abstract}
Online distillation of vision-language-action (VLA) policies commonly relies on teacher action scores that an inference-only interface may not expose. Trajectory-only Adversarial Distillation (\gad{}) instead turns a fixed bank of teacher joint-position features into adaptive feedback: a discriminator compares feature marginals from teacher and student rollouts and supplies rewards for on-policy learning. The student interacts with the environment without online teacher queries or environment task rewards. On ManiSkill \texttt{PutOnPlateInScene25Main}, \gad{} improves OpenVLA-OFT (7B) from $26.2\%$ to $76.6\%$ \successonce{} ($+50.4$ percentage points) and from $20.3\%$ to $58.6\%$ \successend{}. It trails white-box on-policy distillation (OPD) at $81.6\%$ / $59.4\%$ by $5.0$ / $0.8$ percentage points ($n{=}256$ evaluation trajectories per checkpoint). The tested four-epoch continuation achieves the highest \gad{} result without additional teacher data. This best run uses $5\times$ OPD's student interaction budget and $12.5\times$ its complete-rollout batch-passes. The comparison demonstrates online VLA improvement without teacher scoring while quantifying unequal interaction and optimization budgets.
\end{abstract}

\section{Introduction}
\label{sec:intro}

Vision-language-action (VLA) models translate visual observations and language instructions into robot actions, allowing pretrained representations to support manipulation across tasks. OpenVLA and OpenVLA-OFT illustrate the potential of this approach and of supervised fine-tuning (SFT) for adapting a large policy to robot demonstrations~\cite{kim2024openvla,kim2025openvlaoft}. Demonstration-based training nevertheless exposes the policy primarily to states visited by the data-collection policy. At deployment, small action errors can change subsequent observations and move execution away from that distribution. Improving behavior on the student's own visited states is therefore an important complement to fitting demonstration actions.

On-policy distillation (OPD) addresses this mismatch by obtaining teacher supervision on student-generated behavior~\cite{agarwal2024gkd,opd2025}. In the score-based setting considered here, a fixed teacher evaluates the likelihood of action components selected by the student, supplying a direction for online optimization. This access is stronger than the ability to request a demonstration or execute a teacher policy. An inference-only API or proprietary robot stack may return actions without exposing their likelihoods, model weights, or internal representations. Consequently, a capable teacher may be available for collecting successful behavior but unavailable for scoring arbitrary student actions during training.

We separate two resources that are often coupled in online distillation: student environment interaction and continuing teacher access. Our setting permits the student to generate new rollouts throughout optimization, but permits teacher supervision only through an initially collected trajectory bank. The bank remains fixed even as the student changes. This restriction is not offline learning: the student still encounters the consequences of its own actions and can improve through closed-loop interaction. Nor does it provide corrective teacher labels at the student's newly visited states.

\emph{Trajectory-only Adversarial Distillation} (\gad{}) implements this setting with a state-feature discriminator. Teacher rollouts provide a fixed bank of normalized eight-dimensional joint-position features; recent student rollouts provide the comparison distribution. The discriminator produces dense logit rewards for group-relative policy optimization in RLinf~\cite{shao2024deepseekmath,rlinf2025}. Its feedback adapts to the student although the teacher data do not change. The VLA policy itself receives visual-language input, while proprioception supplies the discriminator features. The method matches feature marginals rather than complete temporal trajectories, and the main and best runs use no environment task reward.

A central empirical diagnostic is the distinction between discriminating behavior and improving it. Early configurations separate teacher and student features accurately without raising deterministic task success. Subsequent diagnostics examine discriminator inputs, reward construction, and actor optimization; continuations from a shared checkpoint examine regularization, update epochs, and reward mixing. The tested four-epoch continuation achieves the highest \gad{} evaluation with the unchanged teacher bank. These observations motivate examining optimization alongside reward design, rather than interpreting either discriminator accuracy or a sampled training plateau as a direct measure of the supervision's usefulness.

On OpenVLA-OFT (7B) and ManiSkill, \gad{} raises \successonce{} from $26.2\%$ to $76.6\%$ and \successend{} from $20.3\%$ to $58.6\%$, compared with OPD's $81.6\%$ / $59.4\%$. The comparison shares student initialization, teacher, task, and actor learning rate, but not the training objective or budget. The best \gad{} run uses $5\times$ as many student rollout trajectories and $12.5\times$ as many complete-rollout optimization batch-passes. These results quantify the trade-off between teacher access and student-side training effort.

The paper develops three contributions:
\begin{enumerate}[leftmargin=1.5em,itemsep=2pt]
\item A trajectory-only distillation method coupling a fixed teacher feature bank, adaptive state rewards, and online optimization of a large VLA.
\item Training diagnostics and continuation ablations that distinguish classification quality, sampled rollout behavior, and deterministic policy evaluation.
\item A quantified comparison with white-box OPD, reporting task performance together with the unequal interaction and optimization budgets required by the tested runs.
\end{enumerate}

\section{Related Work}
\label{sec:related}

\subsection{VLA learning and online distillation}

OpenVLA~\cite{kim2024openvla} and OpenVLA-OFT~\cite{kim2025openvlaoft} provide the model family underlying our experiments. Supervised adaptation transfers demonstration behavior into a visual-language policy, but its training distribution need not cover the states induced by the deployed student. DAgger~\cite{ross2011dagger} addresses this issue by gathering expert action labels on learner-visited states and aggregating them into the training set. Its corrective-label access differs from ours: \gad{} collects teacher trajectories before student optimization and makes no subsequent queries at student states. Student interaction alone does not imply access to an expert who can correct each rollout.

Distillation methods also differ in the supervision they request. Knowledge distillation~\cite{hinton2015distilling} transfers teacher output distributions, whereas sequence-level distillation (SeqKD)~\cite{kim2016seqkd} trains on teacher-generated sequences. Generalized knowledge distillation (GKD)~\cite{agarwal2024gkd} evaluates teacher distributions on student-generated samples, bringing supervision closer to the student's current behavior. VLA-OPD~\cite{opd2025} applies on-policy teacher scoring to embodied policies and supplies our white-box comparison. We retain online student rollouts but replace likelihood-based teacher feedback with a discriminator learned from a fixed feature bank. This is a change in supervision access and actor objective, not simply a different estimator of the same OPD loss.

\subsection{Adversarial imitation and observation-based supervision}

GAIL~\cite{ho2016gail} formulates imitation through adversarial discrimination between expert and learner behavior. AIRL~\cite{fu2018airl} develops an adversarial reward-learning formulation, while DAC~\cite{kostrikov2019dac} studies sample efficiency and reward bias with off-policy learning. These works motivate learning a reward from demonstrations rather than specifying a task reward by hand. \gad{} uses a logit transformation of discriminator probabilities, but not AIRL's full reward-recovery architecture. Its discriminator replay should likewise not be confused with off-policy actor learning: the actor updates from the current rollout batch with fixed rewards.

Observation-based imitation removes the need for demonstrated action labels, but does not necessarily remove temporal structure. GAIfO~\cite{torabi2018gaifo} discriminates state transitions, and AMP~\cite{peng2021amp} uses transition pairs to learn motion priors. Our main experiments instead discriminate single-frame joint-position features, so they compare a feature marginal rather than a transition distribution. This narrower representation cannot directly encode object relations or temporal progress. EF-GAIfO~\cite{efgaifo2026} further studies observation-based imitation under embodiment mismatch, including real-robot validation of policies trained in simulation. Our experiments instead retain the same robot embodiment for teacher and student. Design studies of adversarial imitation~\cite{orsini2021ail} motivate examining inputs, reward transformations, and optimization together; gradient-penalty regularization~\cite{gulrajani2017wgan} motivates one continuation ablation. The distinction between a learned proxy and the intended task objective~\cite{skalse2022reward} is especially relevant when discriminator accuracy and task success diverge.

\subsection{Black-box generative distillation}

Black-box language-model distillation offers related ways to replace unavailable teacher likelihoods. GAD~\cite{gad2024llm} trains a text discriminator on teacher and student outputs, using student SeqKD warmup and separate discriminator warmup before adversarial online learning. Our student instead starts from SFT on a different demonstration dataset, and our recipe has no separate discriminator warmup. This difference matters when interpreting early discriminator separation: the initial student need not already approximate the teacher bank. ROPD~\cite{ropd2026} derives prompt-specific rubrics from teacher--student contrasts and uses those rubrics to score student outputs. \gad{} does not induce semantic evaluation criteria; it learns a reward directly from observed robot features.

Adversarial objectives also appear in generative-model acceleration. ADD~\cite{sauer2023add} combines adversarial learning with teacher distillation to accelerate diffusion sampling; an adversarial loss alone therefore does not establish trajectory-only or black-box teacher access. AFD~\cite{afd2026} studies black-box adversarial flow distillation for video generation, but its teacher-access and update protocol should not be identified with our fixed-bank embodied setting. These methods share the use of learned comparisons while differing in available supervision and optimization. Our analysis concerns the interaction between feature-based rewards and online control, where student actions change the future observations used for learning.

\subsection{Concurrent embodied adversarial distillation}

FA-OPD~\cite{faopd2026} is closely related embodied work that learns a flow-matching teacher from demonstrations and co-trains it with lightweight MLP students. Its distillation combines adversarial and mean-squared-error objectives, rather than relying on adversarial rewards alone. In contrast, our teacher is pre-existing and supplies a fixed bank before optimization of an OpenVLA-OFT (7B) student. No teacher model is co-trained, and no teacher action scores or additional demonstrations are requested during student training.

The distinction concerns teacher construction, student scale, supervision availability, and the actor loss. Our evaluation focuses on transferring behavior from a fixed external teacher into a large VLA. Score-based OPD provides a reference for measuring the performance and student-side training budgets associated with replacing online teacher scores by a fixed trajectory bank.

\section{Method}
\label{sec:method}

\subsection{Problem formulation}
\label{sec:formulation}

We consider an episodic manipulation process $\mathcal{M} = (\mathcal{O}, \mathcal{A}, P, r^{\mathrm{env}}, \rho_0, H)$. The environment observation $o_t$ includes camera images, a task instruction, and proprioception $s_t$. The student $\pi_\theta(a_t \mid x_t)$ receives only the visual-language input $x_t=x(o_t)$, not joint positions, and produces an action chunk $a_t$. Here $t$ indexes chunk decisions. A trajectory $\tau = (o_0, a_0, \ldots, o_H)$ induces a marginal distribution of features $\phi(o_t)$ at those decisions, where $\phi : \mathcal{O} \to \mathbb{R}^8$ extracts eight joint positions (qpos8) and normalizes them using statistics fitted once on the teacher bank. The same fixed map is applied to teacher and student observations. It excludes object state; we compare feature marginals, not complete temporal trajectories.

The teacher $\pi_T$ supplies trajectories $\mathcal{D}_T = \{\tau^T_i\}$ once, with no logits, densities, gradients, weights, or hidden states. We seek to improve an SFT-initialized student through its own environment interaction. White-box OPD instead evaluates teacher log-likelihoods for the student's action components throughout training. The methods share student initialization, teacher, benchmark, and actor learning rate, but not identical objectives or training settings.

\subsection{Discriminator and reward}
\label{sec:tad}

A discriminator $D_\psi(z)=\operatorname{sigmoid}(f_\psi(z))$, with $z=\phi(o)$, learns to assign high probabilities to teacher features and low probabilities to student features (Figure~\ref{fig:framework}). The baseline binary cross-entropy objective is
\begin{equation}
\mathcal{L}_D(\psi) = -\mathbb{E}_{z \sim \rho_T}\big[\log D_\psi(z)\big]
  - \mathbb{E}_{z \sim \rho_{\mathcal{B}}}\big[\log\big(1 - D_\psi(z)\big)\big],
\label{eq:dloss}
\end{equation}
where $\rho_T$ is the fixed teacher feature distribution and $\rho_{\mathcal{B}}$ is the empirical student feature distribution in replay $\mathcal{B}$. Replay retains the latest 4 student batches to reduce overfitting to one batch. Equation~\eqref{eq:dloss} shows hard-label BCE; training uses label smoothing with $\eta=0.1$, replacing the teacher target by $1-\eta$ and the student target by $\eta$.

We turn each score into a dense scalar AIRL-style logit reward~\cite{fu2018airl}:
\begin{equation}
r_t \;=\; \log D_\psi\!\big(\phi(o_t)\big) \;-\; \log\big(1 - D_\psi\!\big(\phi(o_t)\big)\big),
\label{eq:reward}
\end{equation}
using one feature sample per action chunk. The reward is positive when $D_\psi>1/2$ and negative when $D_\psi<1/2$: it measures learned teacher-likeness, not ground-truth task success. We use this AIRL-style transformation, not AIRL's full reward-recovery architecture. Optional task-reward mixing adds $\alpha\,r^{\mathrm{env}}_t$; the main and best runs use none.

Main results use single-frame qpos; transition features as in AMP/GAIfO~\cite{peng2021amp,torabi2018gaifo} are an extension, not evaluated here.

\subsection{Policy update}
\label{sec:update}

For each group of $G$ rollouts sharing a task, \gad{} sums rewards into episode returns $R_i = \sum_t r_{i,t}$ and applies group-relative normalization~\cite{shao2024deepseekmath}:
\begin{equation}
\hat{A}_i = \frac{R_i - \bar{R}}{\operatorname{std}_{\mathrm{sample}}(R_1,\ldots,R_G)+\delta},
\label{eq:grpo_adv}
\end{equation}
where $\bar{R}=G^{-1}\sum_i R_i$, the sample standard deviation uses denominator $G-1$, and $\delta=10^{-6}$. A rollout's advantage is shared by all its valid action tokens. Here a token denotes a likelihood component of an action chunk, not necessarily an autoregressively generated symbol. With $j$ indexing these components, define $\ell_{\theta,itj}=\log\pi_{\theta,j}(a_{itj}\mid x_{it})$ and $\rho_{itj}=\exp(\ell_{\theta,itj}-\ell_{\mathrm{old},itj})$. The actor minimizes a PPO-style clipped objective~\cite{schulman2017ppo,shao2024deepseekmath}:
\begin{equation}
\mathcal{L}_{\mathrm{pol}}(\theta) = -\mathbb{E}_{i,t,j}\Big[\min\big(\rho_{itj}\hat{A}_i,\; \operatorname{clip}(\rho_{itj}, 1-\varepsilon_{\mathrm{low}}, 1+\varepsilon_{\mathrm{high}})\hat{A}_i\big)\Big],
\label{eq:grpo}
\end{equation}
where the expectation masks invalid tokens and $\ell_{\mathrm{old},itj}$ is evaluated under the rollout policy. The potentially asymmetric clipping bounds limit incentives for large likelihood-ratio changes.

OPD uses a different actor objective, not Eq.~\eqref{eq:grpo}. Its token-level signal is the teacher log-likelihood minus the old student's log-likelihood. Writing $\ell_{T,itj}=\log\pi_{T,j}(a_{itj}\mid x_{it})$, the loss is
\[
\mathcal{L}_{\mathrm{OPD}}(\theta)
= -\mathbb{E}_{i,t,j}\!\left[\ell_{\theta,itj}\,
\operatorname{stopgrad}\!\left(\ell_{T,itj}-\ell_{\mathrm{old},itj}\right)\right],
\]
with invalid tokens masked. This teacher-score signal is neither a chunk-level scalar nor a group-normalized trajectory return.

Algorithm~\ref{alg:tad} describes the distributed reference implementation. The current discriminator scores each rollout before its update. Rewards remain fixed while the actor and discriminator are updated in parallel; the updated discriminator is used for the next rollout batch.

\begin{algorithm}[!htbp]
\caption{\gad{}: trajectory-only adversarial distillation}
\label{alg:tad}
\begin{algorithmic}[1]
\State Collect teacher trajectories $\mathcal{D}_T$ once; fit $\phi$ and cache bank $\mathcal{F}_T=\{\phi(o_t^T)\}$
\State Initialize SFT student $\pi_\theta$, discriminator $D_\psi$, and replay $\mathcal{B}\gets\emptyset$
\For{step $=1,\ldots,N$}
  \State Set $\theta_{\mathrm{old}}\gets\theta$; roll out the student and extract $\{\phi(o_{it})\}$
  \State Score the current batch with $D_\psi$ using Eq.~\eqref{eq:reward}; fix these rewards
  \State Append current features to $\mathcal{B}$; retain the latest 4 batches
  \State Compute returns and advantages with Eq.~\eqref{eq:grpo_adv}
  \State In parallel: update $\theta$ with Eq.~\eqref{eq:grpo} on the current batch;
  \Statex \hspace{\algorithmicindent} update $D_\psi$ with Eq.~\eqref{eq:dloss} on $(\mathcal{F}_T,\mathcal{B})$
\EndFor
\end{algorithmic}
\end{algorithm}

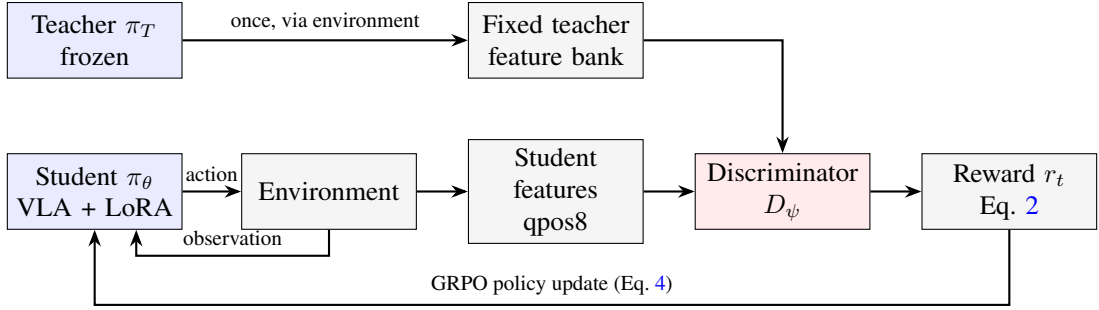
\begin{figure}[!htb]
\centering
\begin{tikzpicture}[
  font=\small,
  box/.style={draw, align=center, text width=2.1cm, minimum height=10mm, inner sep=3pt},
  policy/.style={box, fill=blue!8},
  disc/.style={box, fill=red!8},
  data/.style={box, fill=gray!8},
  arr/.style={-{Stealth[length=2.2mm]}, thick},
]
\node[policy] (T) at (0,2) {Teacher $\pi_T$\\frozen};
\node[data] (DT) at (6.1,2) {Fixed teacher\\feature bank};
\node[policy] (S) at (0,0) {Student $\pi_\theta$\\VLA + LoRA};
\node[data] (E) at (3.1,0) {Environment};
\node[data] (DS) at (6.1,0) {Student features\\qpos8};
\node[disc] (D) at (9.1,0) {Discriminator\\$D_\psi$};
\node[data] (R) at (12.1,0) {Reward $r_t$\\Eq.~\ref{eq:reward}};

\draw[arr] (T) -- node[above,font=\scriptsize]{once, via environment} (DT);
\draw[arr] (DT.east) -- (9.1,2) -- (D.north);
\draw[arr] (S.east) -- node[above,font=\scriptsize]{action} (E.west);
\draw[arr] (E.south) -- (3.1,-0.85) -- node[above,font=\scriptsize]{observation} (0.55,-0.85) -- ([xshift=0.55cm]S.south);
\draw[arr] (E) -- (DS);
\draw[arr] (DS) -- (D);
\draw[arr] (D) -- (R);
\draw[arr] (R.south) -- (12.1,-1.5) -- node[above,font=\scriptsize]{GRPO policy update (Eq.~\ref{eq:grpo})} (0,-1.5) -- (S.south);
\end{tikzpicture}
\caption{Fixed teacher data, online student learning. The top path collects teacher features once. Below, the student acts in the environment, receives observations, and learns from discriminator rewards. The teacher never scores student actions during training.}
\label{fig:framework}
\end{figure}

\FloatBarrier
\section{Experimental Setup}
\label{sec:setup}

\textbf{Task.} We use ManiSkill3~\cite{tao2024maniskill3} \texttt{PutOnPlateInScene25Main-v3}, a Bridge-style pick-and-place task with a WidowX arm and randomized scene composition.

\textbf{Teacher and student.} The student is OpenVLA-OFT (7B), fine-tuned with LoRA~\cite{hu2021lora} via SFT, with baseline \successonce{} $26.2\%$ and \successend{} $20.3\%$. The teacher uses the same architecture after GRPO RL and reaches $93.8\%$ / $77.3\%$ in our local evaluation. The model card separately reports a success rate of $94.1\%$;\footnote{Teacher model card: \url{https://huggingface.co/RLinf/RLinf-OpenVLAOFT-GRPO-ManiSkill3-25ood/raw/main/README.md}.} its metric definition and evaluation protocol have not been established as equivalent to ours. Policy inputs contain images and the instruction, not qpos; the latter is used separately for discriminator features.

\textbf{Metrics and scope.} Standalone evaluation uses greedy decoding on a fixed suite. \successonce{} means success at any point; \successend{} requires success at termination. Main results and continuation ablations use $n{=}256$ trajectories per checkpoint, not training seeds. Differences are point estimates, not evidence of statistical equivalence; training-seed variation is untested. Sampled training rollouts are separate process diagnostics, not interchangeable with deterministic evaluations.

\textbf{OPD (white-box).} The fixed teacher scores individual likelihood components of student action chunks. The actor uses the masked token-level signal $\ell_{T,itj}-\ell_{\mathrm{old},itj}$ in the unclipped OPD objective of \S\ref{sec:update}, with learning rate $5\text{e-5}$. It does not use \gad{}'s group-normalized returns or clipped policy loss.

\textbf{\gad{} (black-box).} A ReLU MLP $[256,256]$ discriminates qpos8 features normalized using the fixed teacher-bank statistics. AdamW trains it with the smoothed BCE described after Eq.~\eqref{eq:dloss}, label smoothing $0.1$, weight decay $0.01$, gradient clip $10$, and the last 4 student batches. The actor uses learning rate $5\text{e-5}$ and one update epoch in the baseline. Both baseline and four-epoch continuation use only Eq.~\eqref{eq:reward}, without environment task reward. Ablations add gradient penalty (\texttt{gp\_coef}) or task reward with $\alpha{=}0.05$.

\textbf{Collection and training units.} One outer iteration (a training ``step'') uses 64 parallel environments and 4 rollout rounds: $64\times4=256$ student trajectories, each with an 80-environment-step horizon and action chunks of length 8. Teacher collection uses the same environment configuration for 5 outer iterations, yielding $12{,}800 = 5 \times 4 \times 64 \times (80/8)$ feature samples at chunk boundaries. This bank is never expanded. Student interaction and optimizer batch-passes are counted separately in \S\ref{sec:budgets}.

\textbf{System.} RLinf~\cite{rlinf2025} provides Ray scheduling, an FSDP actor, and Hydra configuration.

\section{Training Diagnostics and Ablations}
\label{sec:failures}

Accurate discrimination does not guarantee a better policy. The first three diagnostics probe inputs, reward source, and actor updates (Tables~\ref{tab:diag_disc}--\ref{tab:diag_lr}); variant IDs identify configurations, not controlled single-factor comparisons. Several settings, training budgets, and evaluation checkpoints differ. All except v8 use actor lr $\le 1\text{e-5}$. Section~\ref{sec:ablations} then compares continuations from a shared checkpoint.

\subsection{Discriminator design: inputs and training strength}
\label{sec:diag:disc}

With state--action features, discriminator accuracy approached 0.998 but the student did not improve (v1). Weaker discriminator training reduced saturation without raising success (v2; Table~\ref{tab:diag_disc}). Easy separation, potentially through action features, is not enough; neither is avoiding saturation.

\begin{table}[!htb]
\centering
\small
\caption{Discriminator-design diagnostics. Success is standalone deterministic \successonce{}; these variants change several discriminator settings together.}
\label{tab:diag_disc}
\begin{tabular}{@{}c>{\raggedright\arraybackslash}p{5.0cm}c>{\raggedright\arraybackslash}p{4.3cm}@{}}
\toprule
Variant & Configuration & Eval success & Observation \\
\midrule
v1 & $[s_t,a_t]$ 64-dim; GAIL reward $-\log(1\!-\!D)$; 100 $D$ updates/step; no regularization & 25.4\% & Accuracy approaches 0.998 within a few steps. \\
v2 & 8 $D$ updates/step; $D$ lr $1\text{e-4}$; label smoothing 0.1 & $\sim$26\% & Accuracy stays around 0.7--0.8 without a corresponding task gain. \\
\bottomrule
\end{tabular}
\end{table}

State-only qpos8 inputs were used next: v5 stopped at a host out-of-memory failure before evaluation, while v6 also changed reward and regularization. Unlike GAD's teacher-output warmup~\cite{gad2024llm,kim2016seqkd}, our student starts from SFT on a different demonstration dataset; we use no separate discriminator warmup.

\FloatBarrier
\subsection{Reward source and controls}
\label{sec:diag:reward}

Task-reward mixing and additional discriminator updates left evaluation near SFT (v3). Even random-noise rewards produced a sampled rollout peak of $42\%$ (v4), showing why a transient peak is not sufficient evidence of imitation. This control does not show that learned rewards are irrelevant. Combining qpos8, logit rewards, weight decay, and replay reduced discriminator saturation but still did not improve evaluation (v6; Table~\ref{tab:diag_reward}), leaving reward quality and actor optimization as possible bottlenecks.

\begin{table}[!htb]
\centering
\small
\caption{Reward diagnostics. Eval denotes standalone deterministic \successonce{}; v4 has only a sampled rollout observation. v6 changes input, reward, weight decay, and replay together.}
\label{tab:diag_reward}
\begin{tabular}{@{}c>{\raggedright\arraybackslash}p{5.0cm}c>{\raggedright\arraybackslash}p{4.3cm}@{}}
\toprule
Variant & Configuration change & Eval success & Observation \\
\midrule
v3 & v2 + environment reward 0.5; $D$ updates $8\to20$ per step & 26.6\% & No substantial evaluation gain in this run. \\
v4 & Random-noise reward; no discriminator signal & -- & Sampled rollout success peaks at 42\%; no standalone evaluation. \\
v6 & State-only 8-dim qpos; AIRL logit; weight decay 0.01; four-batch replay & 26.2\% & Accuracy is about 0.78, but task evaluation remains near baseline. \\
\bottomrule
\end{tabular}
\end{table}

\FloatBarrier
\subsection{Learning rate and update count}
\label{sec:diag:lr}

At actor lr $2\text{e-6}$, both \gad{} (v6) and a separate OPD run improved little; OPD reached only $28.1\%$ evaluation success. At $1\text{e-5}$, one or four update epochs gave similar sampled rollout levels (v7/v9). At $5\text{e-5}$, v8 reached $68.8\%$ deterministic success at step-150 with the same discriminator recipe (Table~\ref{tab:diag_lr}). These unmatched runs motivate the chosen learning rate, not a causal threshold. The low-rate four-epoch result does not preclude gains at the higher rate tested next.

\begin{table}[!htb]
\centering
\small
\caption{Learning-rate and update-count diagnostics using the v6 discriminator recipe. Reported budgets and checkpoints are not matched; \textsuperscript{*} denotes sampled rollout success rather than standalone evaluation.}
\label{tab:diag_lr}
\begin{tabular}{@{}c>{\raggedright\arraybackslash}p{5.0cm}c>{\raggedright\arraybackslash}p{4.3cm}@{}}
\toprule
Variant & Actor configuration & Success & Observation \\
\midrule
v6 & Actor lr $2\text{e-6}$ & 26.2\% & Low-learning-rate evaluation baseline. \\
v7 & Actor lr $1\text{e-5}$ & $\sim$28\%\textsuperscript{*} & Limited rollout improvement in the observed window. \\
v9 & Actor lr $1\text{e-5}$; four update epochs & $\sim$28\%\textsuperscript{*} & Additional epochs did not improve the observed rollout level. \\
v8 & Actor lr $5\text{e-5}$ & \textbf{68.8\%} & Standalone deterministic evaluation at step-150. \\
\bottomrule
\end{tabular}
\end{table}

\FloatBarrier

\subsection{Ablations}
\label{sec:ablations}

Three branches resume the same baseline step-150 checkpoint, change one setting each, and run to step-300. The baseline also continues to step-300, but is evaluated only at steps 150 and 280: there is no matched baseline step-300 evaluation, so these endpoints do not isolate the effect of update epochs at equal interaction budget. Table~\ref{tab:ablations} preserves these endpoints and the raw rollout ranges. Figure~\ref{fig:ablations} shows nine-point smoothed curves; the baseline range is $59.4\%$--$69.5\%$ over steps 150--300.

\textbf{Discriminator regularization.} The gradient-penalty branch with coefficient 10 does not exceed the step-150 baseline on both evaluation metrics: \successonce{} is close while \successend{} is lower.\footnote{Archived summaries specify coefficient 10 but do not resolve whether discriminator probability or logit was penalized; the exact variant remains unverified.} Its smoothed rollout range is $60.9\%$--$70.1\%$ over the available steps 196--300.

\textbf{Policy updates.} The tested four-epoch continuation achieves the highest \gad{} evaluations, with \successonce{} / \successend{} $7.8$ / $12.1$pp above step-150 without new teacher data. Yet the smoothed rollout range, $59.8\%$--$70.0\%$ over steps 151--300, largely overlaps the baseline: deterministic gains need not produce equally large changes in sampled rollouts.

\textbf{Reward mixing.} The environment-reward branch with coefficient 0.05 has slightly higher \successonce{} than step-150 and unchanged \successend{}, remaining below the four-epoch continuation. Its smoothed rollout range is $58.3\%$--$65.5\%$ over steps 151--300.

\begin{figure}[!htb]
\centering
\includegraphics[width=0.78\columnwidth]{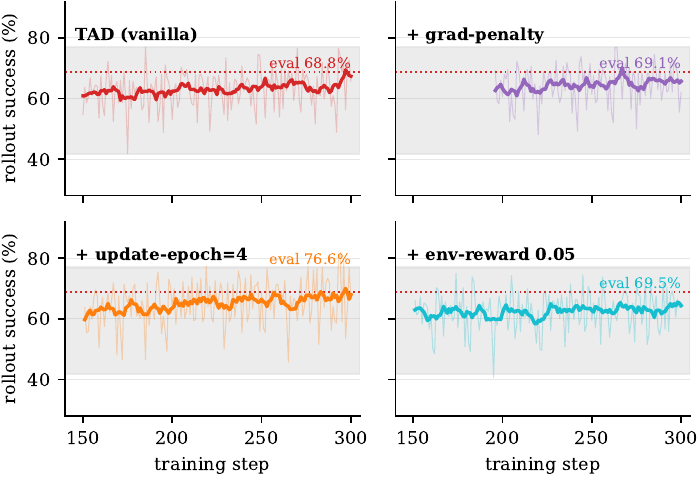}
\caption{Continuations from step-150. Thin lines are raw sampled success; thick lines are centered nine-point moving averages with available-point boundary windows. The gray band is the baseline v8 raw range, not a confidence interval. The red dotted line is the baseline step-150 deterministic \successonce{} ($68.8\%$), not a branch evaluation; annotations give each branch's independent evaluation.}
\label{fig:ablations}
\end{figure}

\begin{table}[!htb]
\centering
\footnotesize
\setlength{\tabcolsep}{4pt}
\caption{Continuation results. Raw ranges cover the full available log intervals listed below. Evaluations use 256 trajectories each; checkpoint labels distinguish the baseline's step-150/280 evaluations from the branches' step-300 evaluations.}
\label{tab:ablations}
\begin{tabular}{@{}>{\raggedright\arraybackslash}p{2.05cm}>{\raggedright\arraybackslash}p{3.9cm}>{\raggedright\arraybackslash}p{1.4cm}>{\raggedright\arraybackslash}p{2.0cm}>{\raggedright\arraybackslash}p{4.2cm}@{}}
\toprule
Configuration & Change & Log steps & Raw success & Eval once / end (checkpoint) \\
\midrule
Baseline v8 & -- & 150--300 & 41.8--77.0\% & 68.8\% / 46.5\% (s150)\newline 66.4\% / 44.1\% (s280) \\
Gradient penalty & \texttt{gp\_coef=10} & 196--300 & 48.0--77.0\% & 69.1\% / 41.0\% (s300) \\
Policy epochs & \texttt{update\_epoch=4} & 151--300 & 45.7--82.0\% & \textbf{76.6\%} / \textbf{58.6\%} (s300) \\
Env-reward mix & \texttt{env\_reward\_coef=0.05} & 151--300 & 40.6--75.0\% & 69.5\% / 46.5\% (s300) \\
\bottomrule
\end{tabular}
\end{table}

\FloatBarrier
\section{Results and Analysis}
\label{sec:results}

\subsection{Black-box versus white-box distillation}
\label{sec:headtohead}

\begin{figure}[!htb]
\centering
\includegraphics[width=0.78\columnwidth]{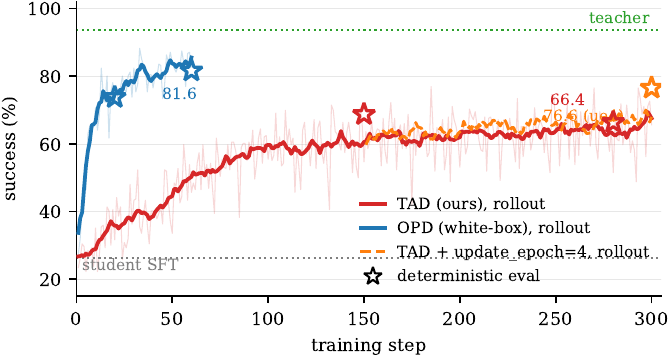}
\caption{Sampled training success versus outer iteration. Thin lines are raw values; centered moving averages use 11 points for baseline \gad{}, 5 for OPD, and 9 for the dashed four-epoch continuation (ue4) from step-150, with available-point boundary windows. Stars mark deterministic evaluations under a different decoding protocol. The ue4 branch reaches $76.6\%$ evaluation success despite overlapping the baseline's sampled rollout range.}
\label{fig:convergence}
\end{figure}

Teacher trajectories alone support a $50.4$pp gain in \successonce{} over SFT. The best \gad{} checkpoint reaches $76.6\%$ / $58.6\%$, versus OPD's $81.6\%$ / $59.4\%$ (Table~\ref{tab:main}). Baseline \gad{} shows no further gain at its later evaluated checkpoint; the four-epoch continuation reaches higher success with the same teacher bank. Figure~\ref{fig:convergence} places these evaluations alongside sampled training curves.

The best \gad{} run reaches approximately $94\%$ of OPD's \successonce{} and $99\%$ of its \successend{}, compared with approximately $84\%$ of OPD's \successonce{} for baseline \gad{} at step-150. These are ratios of raw success rates, not fractions of the improvement over SFT, and the training budgets differ.

\begin{table}[!htb]
\centering
\small
\caption{Standalone deterministic evaluation on ManiSkill \texttt{PutOnPlateInScene25Main}, with 256 trajectories per checkpoint. The four-update-epoch branch resumes from \gad{} step-150 and continues to step-300.}
\label{tab:main}
\begin{tabular}{@{}lcccc@{}}
\toprule
Method & ckpt & \successonce{} & \successend{} & $n$ \\
\midrule
Student (SFT) & -- & 26.2\% & 20.3\% & 256 \\
\midrule
\multirow{2}{*}{\gad{} (black-box)} & step-150 & 68.8\% & 46.5\% & 256 \\
 & step-280 & 66.4\% & 44.1\% & 256 \\
\gad{} + 4$\times$ update epochs & step-300 & \textbf{76.6\%} & \textbf{58.6\%} & 256 \\
\midrule
\multirow{2}{*}{\opd{} (white-box)} & step-20 & 73.8\% & 53.1\% & 256 \\
 & step-60 & \textbf{81.6\%} & \textbf{59.4\%} & 256 \\
\midrule
Teacher & -- & 93.8\% & 77.3\% & 256 \\
\bottomrule
\end{tabular}
\end{table}

\subsection{Training budgets}
\label{sec:budgets}

The best \gad{} run uses 300 iterations: one update epoch for the first 150 and four for the remaining 150. OPD uses 60 iterations with one epoch per batch. Thus \gad{} uses $5\times$ the student rollout trajectories ($300\times256$ versus $60\times256$) and $12.5\times$ the optimizer batch-passes ($(150+150\times4)/60$), with no added teacher data.

A batch-pass is one optimization pass through a \emph{complete rollout batch}, not one minibatch update. These ratios measure neither GPU time, wall-clock time, nor total computation. Likewise, baseline \gad{} at step-150 and OPD at step-20 describe early progress at different success levels, not equal-quality endpoints or final convergence.

\FloatBarrier
\subsection{Reaching and retaining success}
\label{sec:retention}

\begin{figure}[!htb]
\centering
\includegraphics[width=0.78\columnwidth]{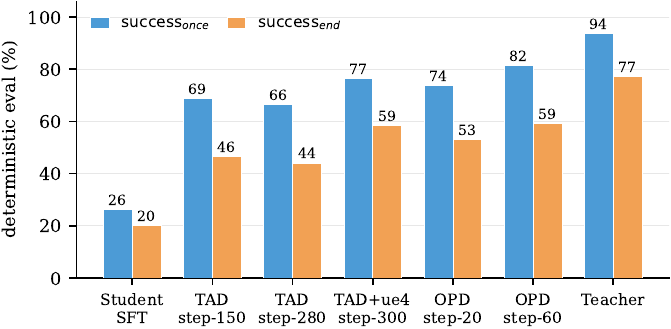}
\caption{Deterministic \successonce{} and \successend{} ($n{=}256$ trajectories each). Once-to-end gaps are 22.3pp for baseline \gad{} at step-150, 22.2pp for OPD at step-60, and 18.0pp for \gad{}+ue4 at step-300.}
\label{fig:once_end}
\end{figure}

Baseline \gad{} at step-150 trails OPD at step-60 by $12.8$pp on \successonce{} and $12.9$pp on \successend{}; the higher-update branch reduces these gaps to $5.0$ / $0.8$pp. Both methods can reach success and then lose it (Figure~\ref{fig:once_end}). This aggregate pattern does not establish a common cause.

\begin{figure}[!htb]
\centering
\includegraphics[width=\columnwidth]{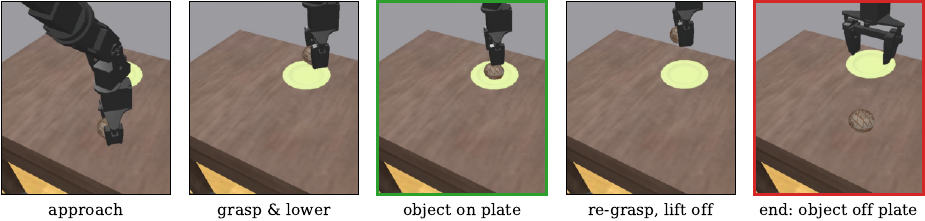}
\caption{One episode from the \gad{} step-150 deterministic evaluation, shown from left to right. The student places the object on the plate (green frame, \successonce{} true), then re-grasps and removes it (red frame, \successend{} false).}
\label{fig:failure}
\end{figure}

One actual \gad{} evaluation episode makes the distinction concrete (Figure~\ref{fig:failure}): the student puts the object on the plate, then grasps it again and removes it. The episode counts as \successonce{}, but not \successend{}. Reaching the goal is not the same as leaving the scene in the desired state.

\FloatBarrier

\subsection{Interpretation and evidence scope}
\label{sec:analysis}

A fixed teacher bank can support online improvement while the student interacts with the environment and the discriminator adapts. The four-epoch continuation reaches higher evaluated success than the baseline's evaluated checkpoints with the same bank, so the baseline plateau is not the highest performance demonstrated under this supervision. Language-model OPD work also examines optimizer effort~\cite{rethinking2_2026}; the changing reward model and continuing environment interaction in \gad{} do not establish the same mechanism.

The comparison concerns the source, granularity, and use of feedback, not scalar versus non-scalar supervision. OPD supplies a teacher-minus-old-student log-likelihood signal at each valid action token and optimizes an unclipped, stop-gradient-weighted loss. \gad{} instead sums one discriminator reward per action chunk into a trajectory return, normalizes returns within a group, and shares the resulting advantage across valid action tokens in a clipped objective. Its reward model also changes during training. These differences may affect optimization, but the budget counts do not measure signal information content or identify a single cause of the performance gap.

The chosen feature representation also limits what the reward can express. qpos8 omits object pose, gripper--object relations, and temporal progress. Similar feature marginals can therefore accompany different task outcomes. Richer spatial or temporal features are candidates for future study, not tested explanations for the remaining gap. State-only discrimination is a design choice, not a requirement of trajectory-only teacher access.

Evidence covers one manipulation task and one student architecture. Without behavioral cloning on the same teacher trajectories, we establish improvement over SFT and a quantified comparison with OPD, not superiority over BC. Neither method's best attainable performance is determined. Archived summaries and curves do not substitute for complete per-trajectory evaluation logs and frozen executed-configuration snapshots, which are not currently available; numerical results are retained as reported.

\section{Conclusion}
\label{sec:conclusion}

\gad{} enables online VLA improvement using a fixed bank of teacher joint-position features rather than teacher action scores. An adaptive discriminator and on-policy student updates produce substantial gains without environment task rewards. The tested four-epoch continuation achieves the highest \gad{} result with unchanged teacher data, while using more student interaction and optimization than the reported OPD endpoint. The results demonstrate the feasibility of this restricted teacher-access setting and quantify its performance--budget trade-off on the evaluated manipulation task.


\begin{thebibliography}{99}\small

\bibitem{ho2016gail}
J.~Ho and S.~Ermon.
\newblock Generative adversarial imitation learning.
\newblock \emph{NeurIPS}, 2016.

\bibitem{fu2018airl}
J.~Fu, K.~Luo, and S.~Levine.
\newblock Learning robust rewards with adversarial inverse reinforcement learning.
\newblock \emph{ICLR}, 2018.

\bibitem{kostrikov2019dac}
I.~Kostrikov et~al.
\newblock Discriminator-actor-critic: Addressing sample inefficiency and reward bias in adversarial imitation learning.
\newblock \emph{ICLR}, 2019.

\bibitem{peng2021amp}
X.~B.~Peng et~al.
\newblock AMP: Adversarial motion priors for stylized physics-based character control.
\newblock \emph{SIGGRAPH}, 2021.

\bibitem{orsini2021ail}
M.~Orsini et~al.
\newblock What matters for adversarial imitation learning?
\newblock \emph{NeurIPS}, 2021.

\bibitem{gulrajani2017wgan}
I.~Gulrajani et~al.
\newblock Improved training of Wasserstein GANs.
\newblock \emph{NeurIPS}, 2017.

\bibitem{skalse2022reward}
J.~Skalse et~al.
\newblock Defining and characterizing reward gaming.
\newblock \emph{NeurIPS}, 2022.

\bibitem{hinton2015distilling}
G.~Hinton, O.~Vinyals, and J.~Dean.
\newblock Distilling the knowledge in a neural network.
\newblock arXiv:1503.02531, 2015.

\bibitem{agarwal2024gkd}
R.~Agarwal et~al.
\newblock On-policy distillation of language models: Learning from self-generated mistakes.
\newblock \emph{ICLR}, 2024.

\bibitem{kim2016seqkd}
Y.~Kim and A.~M.~Rush.
\newblock Sequence-level knowledge distillation.
\newblock \emph{EMNLP}, 2016.

\bibitem{kim2024openvla}
M.~J.~Kim et~al.
\newblock OpenVLA: An open-source vision-language-action model.
\newblock \emph{CoRL}, 2024.

\bibitem{kim2025openvlaoft}
M.~J.~Kim, C.~Finn, and P.~Liang.
\newblock Fine-tuning vision-language-action models: Optimizing speed and success.
\newblock \emph{RSS}, 2025.

\bibitem{opd2025}
Z.~Zhong et~al.
\newblock VLA-OPD: Bridging offline SFT and online RL for vision-language-action models via on-policy distillation.
\newblock arXiv:2603.26666, 2026.

\bibitem{faopd2026}
Z.~Wan et~al.
\newblock Adversarial dual on-policy distillation from expressive teacher.
\newblock arXiv:2605.27095, 2026.

\bibitem{sauer2023add}
A.~Sauer et~al.
\newblock Adversarial diffusion distillation.
\newblock \emph{ECCV}, 2024.

\bibitem{gad2024llm}
T.~Ye et~al.
\newblock Black-box on-policy distillation of large language models.
\newblock arXiv:2511.10643, 2025.

\bibitem{ross2011dagger}
S.~Ross, G.~Gordon, and D.~Bagnell.
\newblock A reduction of imitation learning and structured prediction to no-regret online learning.
\newblock \emph{AISTATS}, 2011.

\bibitem{torabi2018gaifo}
F.~Torabi, G.~Warnell, and P.~Stone.
\newblock Generative adversarial imitation from observation.
\newblock arXiv:1807.06158, 2018.

\bibitem{efgaifo2026}
Y.~Takebayashi et~al.
\newblock Experience-based feasibility-aware generative adversarial imitation from observation under embodiment mismatch.
\newblock arXiv:2610.01171, 2026.

\bibitem{ropd2026}
J.~Fang et~al.
\newblock Rubric-based on-policy distillation.
\newblock arXiv:2605.07396, 2026.

\bibitem{afd2026}
Y.~Luo et~al.
\newblock On-policy adversarial flow distillation for autoregressive video generation.
\newblock arXiv:2605.26105, 2026.

\bibitem{rethinking2_2026}
Z.~Fu et~al.
\newblock Rethinking on-policy distillation of large language models II: One training example.
\newblock arXiv:2609.04172, 2026.

\bibitem{rlinf2025}
C.~Yu et~al.
\newblock RLinf: Flexible and efficient large-scale reinforcement learning via macro-to-micro flow transformation.
\newblock \emph{OSDI}, 2026.

\bibitem{schulman2017ppo}
J.~Schulman et~al.
\newblock Proximal policy optimization algorithms.
\newblock arXiv:1707.06347, 2017.

\bibitem{shao2024deepseekmath}
Z.~Shao et~al.
\newblock DeepSeekMath: Pushing the limits of mathematical reasoning in open language models.
\newblock arXiv:2402.03300, 2024.

\bibitem{tao2024maniskill3}
S.~Tao et~al.
\newblock ManiSkill3: GPU parallelized robotics simulation and rendering for generalizable embodied AI.
\newblock arXiv:2410.00425, 2024.

\bibitem{hu2021lora}
E.~J.~Hu et~al.
\newblock LoRA: Low-rank adaptation of large language models.
\newblock arXiv:2106.09685, 2021.

\end{thebibliography}
\end{document}